\pdfoutput=1

\documentclass[11pt]{article}

\usepackage{acl}

\usepackage{times}
\usepackage{latexsym}
\usepackage{multirow}
\usepackage{tikz}
\usetikzlibrary{positioning, shapes}
\usepackage{algorithm}
\usepackage{algorithmic}
\usepackage{xcolor}
\usepackage{soul}
\usepackage{authblk}

\usepackage[T1]{fontenc}

\usepackage[utf8]{inputenc}

\usepackage{microtype}

\usepackage{inconsolata}
\usepackage{graphicx}
\usepackage{amsmath}
\usepackage{amssymb}
\usepackage{booktabs}
\usepackage{colortbl}
\usepackage{xspace}
\definecolor{lightgray}{gray}{0.9}

\usepackage{tcolorbox}

\newcommand{\method}{\textsc{ReMALIS}\xspace}

\title{Towards Collaborative Intelligence: Propagating Intentions and Reasoning for Multi-Agent Coordination with Large Language Models}

\author[$^{\diamondsuit \dag}$]{Xihe Qiu}
\author[$^{\diamondsuit \dag}$]{Haoyu Wang}
\author[$^{*\heartsuit \dag}$]{Xiaoyu Tan}
\author[$^{\heartsuit}$]{Chao Qu}
\author[$^\diamondsuit$]{Yujie Xiong}
\author[$^\spadesuit$]{Yuan Cheng}
\author[$^\spadesuit$]{Yinghui Xu}

\author[$^{\heartsuit}$]{Wei Chu}
\author[$^\spadesuit$]{Yuan Qi}
\affil[$^{\diamondsuit}$]{Shanghai University of Engineering Science, Shanghai, China}
\affil[$^{\heartsuit}$]{INF Technology(Shanghai) Co., Ltd., Shanghai, China}
\affil[$^\spadesuit$]{Fudan University, Shanghai, China}
\affil[*]{This is to indicate the corresponding author}
\affil[$\dag$]{This is to indicate the equal contribution}

\begin{document}
\maketitle
\begin{abstract}
Effective collaboration in multi-agent systems requires communicating goals and intentions between agents. Current agent frameworks often suffer from dependencies on single-agent execution and lack robust inter-module communication, frequently leading to suboptimal multi-agent reinforcement learning (MARL) policies and inadequate task coordination. 
To address these challenges, we present a framework for training large language models (LLMs) as collaborative agents to enable coordinated behaviors in cooperative MARL. Each agent maintains a private intention consisting of its current goal and associated sub-tasks. Agents broadcast their intentions periodically, allowing other agents to infer coordination tasks. A propagation network transforms broadcast intentions into teammate-specific communication messages, sharing relevant goals with designated teammates. The architecture of our framework is structured into planning, grounding, and execution modules.  During execution, multiple agents interact in a downstream environment and communicate intentions to enable coordinated behaviors. The grounding module dynamically adapts comprehension strategies based on emerging coordination patterns, while feedback from execution agents influnces the planning module, enabling the dynamic re-planning of sub-tasks. Results in collaborative environment simulation demonstrate intention propagation reduces miscoordination errors by aligning sub-task dependencies between agents. Agents learn when to communicate intentions and which teammates require task details, resulting in emergent coordinated behaviors. This demonstrates the efficacy of intention sharing for cooperative multi-agent RL based on LLMs.
\end{abstract}

\section{Introduction}
With the recent advancements of large language models (LLMs), developing intelligent agents that can perform complex reasoning and long-horizon planning has attracted increasing research attention \cite{sharan2023llmassist,huang2022parallel}. A variety of agent frameworks have been proposed, such as ReAct \cite{yao2022react}, LUMOS \cite{yin2023lumos}, Chameleon \cite{lu2023chameleon} and BOLT \cite{chiu2024computational}. These frameworks typically consist of modules for high-level planning, grounding plans into executable actions, and interacting with environments or tools to execute actions \cite{rana2023sayplan}.


Despite their initial success, existing agent frameworks may experience some limitations. Firstly, most of them rely on a single agent for execution \cite{song2023llmplanner,hartmann2022long}. However, as tasks become more complex, the action dimension can be increased exponentially, and it poses significant challenges for a single agent to handle all execution functionalities \cite{chebotar2023qtransformer,wen2023empowering}. Secondly, existing frameworks lack inter-module communication mechanisms. Typically, the execution results are directly used as input in the planning module without further analysis or coordination \cite{zeng2023largelanguage,wang2024subequivariant}. When execution failures occur, the agent may fail to adjust its strategies accordingly \cite{chaka2023generative}. 
Thirdly, the grounding module in existing frameworks operates statically, without interactions with downstream modules. It grounds plans independently without considering feedback or states of the execution module \cite{xi2023rise}. LLMs struggle to handle emergent coordination behaviors and lack common grounding on shared tasks. Moreover, existing multi-agent reinforcement learning (MARL) methods often converge on suboptimal policies that fail to exhibit a certain level of cooperation \cite{gao2023retrievalaugmented,yu2023llm}.

\textit{How can the agents with LLMs effectively communicate and collaborate with each other?} we propose a novel approach, \textbf{Re}cursive \textbf{M}ulti-\textbf{A}gent \textbf{L}earning with \textbf{I}ntention \textbf{S}haring (\method\footnote{The code can be accessed at the following URL:\url{https://github.com/AnonymousBoy123/ReMALIS}.}) to address the limitations of existing cooperative artificial intelligence (AI) multi-agent frameworks with LLMs. \method employs intention propagation between LLM agents to enable a shared understanding of goals and tasks. This common grounding allows agents to align intentions and reduce miscoordination. Additionally, we introduce bidirectional feedback loops between downstream execution agents and upstream planning and grounding modules. This enables execution coordination patterns to guide adjustments in grounding strategies and planning policies, resulting in more flexible emergent behaviors \cite{topsakal2023creating}. By integrating these mechanisms, \method significantly improves the contextual reasoning and adaptive learning capabilities of LLM agents during complex collaborative tasks. The execution module utilizes specialized agents that collaboratively execute actions, exchange information, and propagate intentions via intention networks. These propagated intentions reduce miscoordination errors and guide grounding module adjustments to enhance LLM comprehension based on coordination patterns \cite{dong2023codescore}. Furthermore, execution agents can provide feedback to prompt collaborative re-planning in the planning module when necessary.

Compared to single-agent frameworks, the synergistic work of multiple specialized agents enhances \method's collective intelligence and leads to emerging team-level behaviors \cite{wang2023survey}. The collaborative design allows for dealing with more complex tasks that require distributed knowledge and skills. We demonstrate that: 
\begin{itemize}
    \item Intention propagation between execution agents enables emergent coordination behaviors and reduces misaligned sub-tasks.
    \item Grounding module strategies adjusted by intention sharing improve LLM scene comprehension.
    \item Planning module re-planning guided by execution feedback increases goal-oriented coordination.
\end{itemize} 


Compared to various single-agent baselines and existing state-of-the-art MARL \cite{hu2023language,zou2023wireless} methods using LLMs, our \method framework demonstrates improved performance on complex collaborative tasks, utilizing the publicly available large-scale traffic flow prediction (TFP) dataset and web-based activities dataset. This demonstrates its effectiveness in deploying LLMs as collaborative agents capable of intention communication, strategic adjustments, and collaborative re-planning \cite{du2023review}.

\section{Preliminary}
In this section, we introduce the methods of the proposed \method framework in detail. As illustrated in Figure \ref{total}, \method consists of four key components: 
\begin{figure*}
\centering
\includegraphics[width=0.995\textwidth]{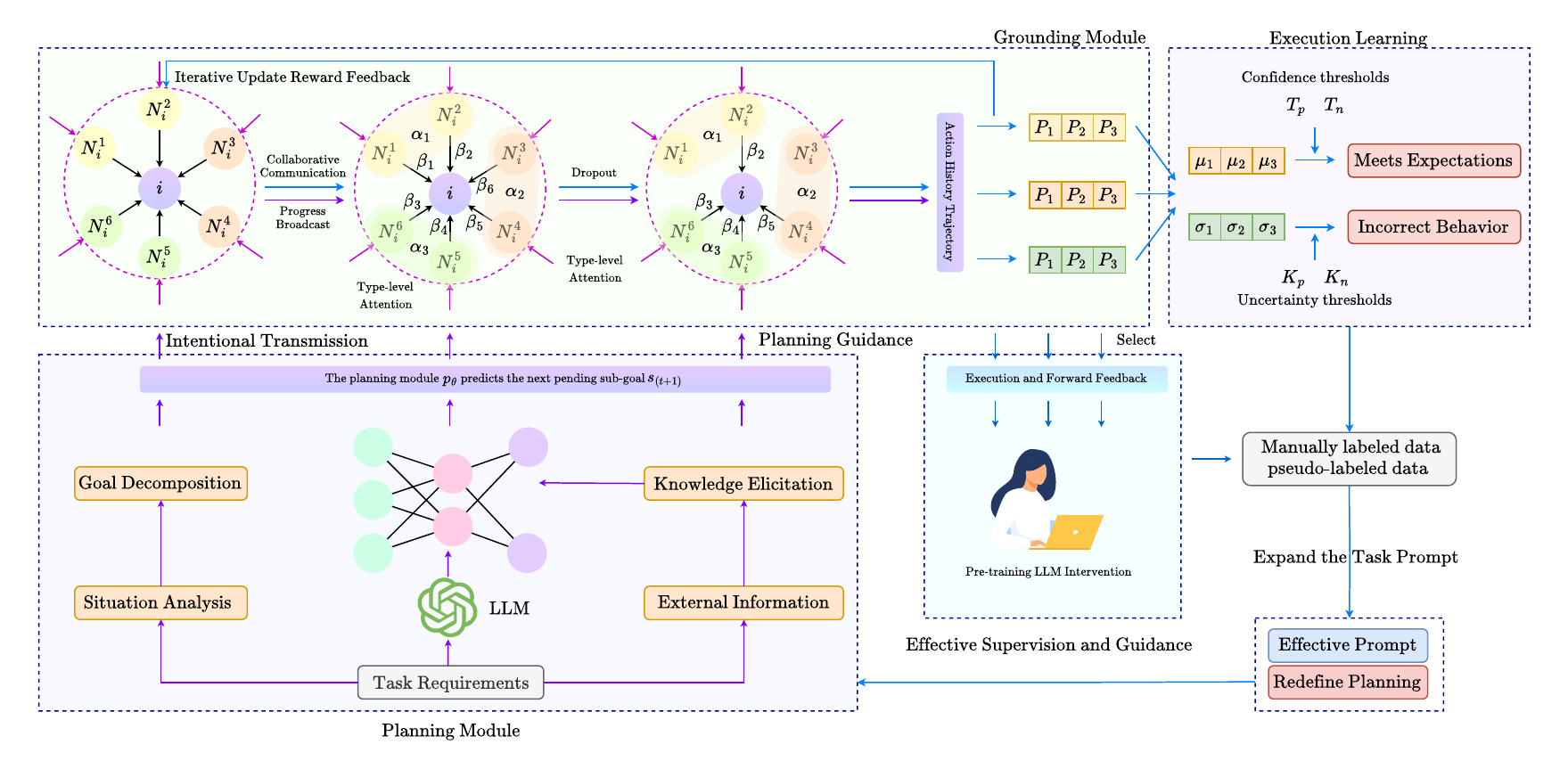}  
\vspace{-5ex}
    \caption{This framework introduces a multi-agent learning strategy designed to enhance the capabilities of LLMs through cooperative coordination. It enables agents to collaborate and share intentions for effective coordination, and utilizes recursive reasoning to model and adapt to each other's strategies.}
    \label{total}
    \vspace{-3ex}
\end{figure*}

\textbf{Planning Module} $p_\theta$ predicts the next pending sub-goal $s_{t+1}$, given the current sub-goal $s_t$ and other inputs
$
s_{t+1} = p_\theta(s_t, I_t, e_t, f_t),
$ where $I_t$ is the current intention, $e_t$ is the grounded embedding, and $f_t$ is agent feedback. $p_\theta$ first encode information through encoding layers $h_t = Encoder(s_t, I_t, e_t, f_t)$ and subsequently predict the sub-goal through $s_{t+1} = Softmax(T_\theta(h_t))$, where $T_\theta$ utilizes the graph neural network (GNN) architecture.

The module is trained to maximize the likelihood of all sub-goals along the decision sequences given the current information on time step $t$. This allows the dynamic re-planning of sub-task dependencies based on agent feedback.
\vspace{-1ex}
\begin{equation}
\theta^* = \arg\max_\theta \prod_{t=1}^T p_\theta(s_{t+1} | s_t, I_t, e_t, f_t).
\end{equation}

\textbf{Grounding Module} $g_\phi$ contextualizes symbol embeddings $e_t = g_\phi(s_t, I_t, f_{1:t})$, where $s_t$, $I_t$, and $f_{1:t}$ represent the states, intention, and feedback up to time step $t$, respectively. These embeddings are processed by encoders $h_t = \text{Encoder}(s_t, I_t, f_{1:t})$ and then by cross-attention layers and convolutional feature extractors: $e_t = Conv(Attn(h_t, V)) + P_t$ over vocabulary $V$. Here, $P_t$ includes agent feedback to enhance grounding accuracy based on coordination signals for more accurate contextual understanding. The module maps language symbols to physical environment representations through:
\begin{equation}
g(x) = f_\theta \left( \sum_{i=1}^N w_i g(x_i) \right),
\end{equation}
where $g(x)$ is the grounded embeddings of policy set $x$ and $g(x_i)$ represents its individual action embedding on agent $i$, respectively, and $w_i$ are learnable weights. The grounding function $f_\theta$ utilizes a GNN architecture for structural composition. Additionally, we employ an uncertainty modeling module that represents ambiguities in grounding:
\begin{equation}
q_\phi(z|x) = \text{Normal}\big(z; \mu_\phi(x), \sigma^2_\phi(x)\big),
\end{equation}
where $z$ is a latent variable modeled as a normal distribution, enabling the capture of multimodal uncertainties in grounding.

\textbf{Cooperative Execution Module} comprises $N$ specialized agents $\{A_1, ..., A_N\}$. This architecture avoids using a single agent to handle all tasks. Instead, each agent is dedicated to a distinct semantic domain, cultivating expertise specific to that domain. For instance, agents $A_1, A_2, $ and $A_3$ may be dedicated to query processing, information retrieval, and arithmetic operations, respectively. This specialization promotes an efficient distribution of tasks and reduces overlap in capabilities.


Decomposing skills into specialized agents risks creating isolated capabilities that lack coordination. To address this, it is essential that agents not only excel individually but also comprehend the capacities and limitations of their peers. We propose an integrated training approach where specialized agents are trained simultaneously to foster collaboration and collective intelligence. We represent the parameters of agent $A_i$ as $\theta_i$. Each agent’s policy, denoted as $y_i \sim \pi_{\theta_i}(\cdot|s)$, samples an output $y_i$ from a given input state $s$. The training objective for our system is defined by the following equation:
\begin{equation}
L_{exe} = \sum_{i=1}^{N}\mathbb{E}_{(s, y^\star)\sim \mathcal D}{\ell(\pi_{\theta_i}(y_i|s), y^\star)},
\end{equation} where $\ell(\cdot)$ represents the task-specific loss function, comparing the agent-generated output $y_i$ with the ground-truth label $y^\star$. $\mathcal D$ denotes the distribution of training data. By optimizing this objective collectively across all agents, each agent not only improves its own output accuracy but also enhances the overall team's ability to produce coherent and well-coordinated results.

During training, we adjust the decomposition of grounding tasks to enhance collaboration, which is represented by the soft module weights $\{w_1, ..., w_N\}$. These weights indicate how the distribution of grounding commands can be optimized to better utilize the capabilities of different agents. The objective of this training is defined by the following loss function:
$L_{com} = \ell(d, w^\star)$, where $\ell$ represents the loss function, $d$ is expressed as subgoal task instruction data, and $w^\star$ signifies the optimal set of weights.

\begin{figure*}
\centering
\includegraphics[width=0.95\textwidth]{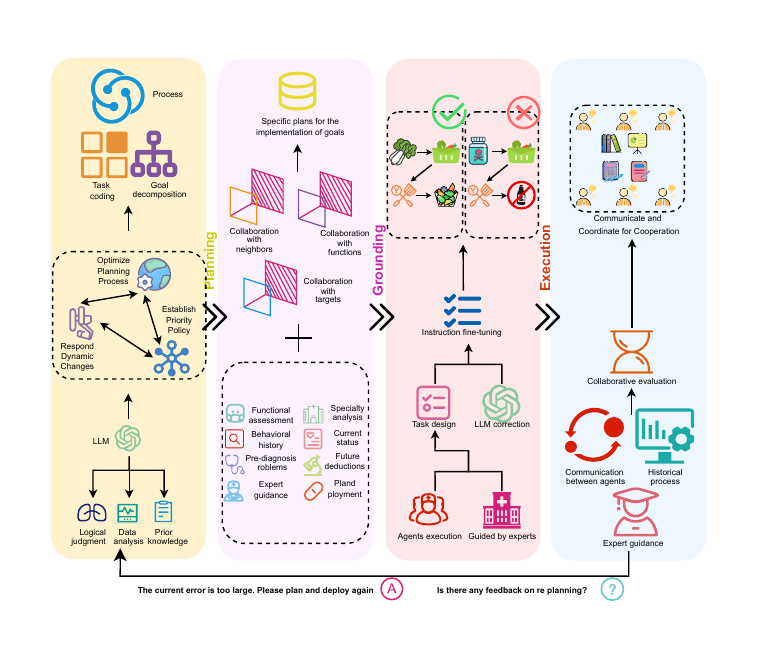}
\vspace{-9ex}
\caption{Overview of the proposed \method: This framework comprises a planning module, grounding module, cooperative execution module, and intention coordination channels.}
\label{P5}
\vspace{-3ex}
\end{figure*}

\section{Approach}
The collaborative MARL of \method focuses on three key points: intention propagation for grounding, bidirectional coordination channels, and integration with recursive reasoning agents. Detailed parameter supplements and pseudocode details can be found in Appendix \ref{ss1} and Appendix \ref{ss2}.

\subsection{Planning with Intention Propagation}
We formulate a decentralized, partially observable Markov game for multi-agent collaboration. Each agent $i$ maintains a private intention $\mathcal{I}_i$ encoded as a tuple $\mathcal{I}_i = (\gamma_i, \Sigma_i, \pi_i, \delta_i)$, where $\gamma_i$ is the current goal, $\Sigma_i = \{\sigma_{i1}, \sigma_{i2}, \ldots\}$ is a set of related sub-goals, $\pi_i(\sigma)$ is a probability distribution over possible next sub-goals, and $\delta_i(\sigma)$ is the desired teammate assignment for sub-goal $\sigma$.

Intentions are propagated through a communication channel $f_\Lambda$ parameterized by $\Lambda$. For a received message $m_{ij}$ from agent $j$, agent $i$ infers a belief over teammate $j$'s intention $b_i(\mathcal{I}_j | m_{ij}) = f_\Lambda(m_{ij})$, where $\Lambda$ is a recurrent neural network. The channel $f_\theta$ is trained in an end-to-end manner to maximize the coordination reward function $R_c$. This propagates relevant sub-task dependencies to enhance common grounding on collaborative goals.
\begin{equation}
\Lambda^* = \arg\max_\Lambda \mathbb{E}_{\mathcal{I}, m \sim f_\Lambda}[R_c(\mathcal{I}, m)].
\end{equation} At each time-step $t$, the LLM witll processinputs comprising the agent's state $s_t$, the intention $\mathcal{I}_t$, and the feedback $f_{1:t}$.

\subsection{Grounding with Bidirectional Coordination Channels}
The execution agent policies, denoted by $\pi_\xi(a_i | s_i, \mathcal{I}_i)$, are parameterized by $\xi$ and conditioned on the agent's state $s_i$ and intention $\mathcal{I}_i$. Emergent coordination patterns are encoded in a summary statistic $c_t$ and passed to upstream modules to guide planning and grounding adjustments. For example, frequent miscoordination on sub-goal $\sigma$ indicates the necessity to re-plan $\sigma$ dependencies in $\mathcal{I}$.

This bidirectional feedback aligns low-level execution with high-level comprehension strategies. In addition to the downstream propagation of intents, execution layers provide bidirectional feedback signals $\psi(t)$ to upstream modules $\psi(t) = \Phi(h^\text{exec}_t)$:
\begin{equation}
h^\text{exec}_t = [\phi_1(o_1), \ldots, \phi_N(o_N)],
\end{equation} where $\Phi(\cdot)$ aggregates agent encodings to summarize emergent coordination, and $\phi_i(\cdot)$ encodes the observation $o_i$ for agent $i$.

Execution agents generate feedback $f_t$ to guide upstream LLM modules through: $f_t = g_\theta(\tau_{1:t})$, where $g_\theta$ processes the action-observation history $\tau_{1:t}$. These signals include coordination errors $\mathcal{E}_t$ which indicate misalignment of sub-tasks; grounding uncertainty $\mathcal{U}_t$, measured as entropy over grounded symbol embeddings; and re-planning triggers $\mathcal{R}_t$, which flag the need for sub-task reordering. These signals can reflect inconsistencies between sub-task objectives, the ambiguity of symbols in different contexts, and the need to adjust previous sub-task sequencing.
\begin{algorithm}[H]
\caption{\method: Recursive Multi-Agent Learning with Intention Sharing}
\begin{algorithmic}[1]
\STATE Initialize LLM parameters $\theta, \phi, \omega$
\STATE Initialize agent policies $\pi_\xi$, communication channel $f_\theta$
\STATE Initialize grounding confusion matrix $C$, memory $M$
\FOR{each episode}
\FOR{each time step $t$}
\STATE Observe states $s_t$ and feedback $f_{1:t}$ for all agents
\STATE Infer intentions $\mathcal{I}_t$ from $s_t, f_{1:t}$ using $\text{LLM}_\theta$
\STATE Propagate intentions $\mathcal{I}_t$ through channel $f_\theta$
\STATE Compute grounded embeddings $e_t = g_\phi(s_t, \mathcal{I}_t, f_{1:t})$
\STATE Predict sub-tasks $\Sigma_{t+1} = p_\theta(\mathcal{I}_t, e_t, f_{1:t})$
\STATE Generate actions $a_t = a_\omega(e_t, \Sigma_{t+1}, f_{1:t})$
\STATE Execute actions $a_t$ and observe rewards $r_t$, new states $s_{t+1}$
\STATE Encode coordination patterns $c_t = \Phi(h^\text{exec}_t)$
\STATE Update grounding confusion $C_t, M_t$ using $c_t$
\STATE Update policies $\pi_\xi$ using $R$ and auxiliary loss $\mathcal{L}_\text{aux}$
\STATE Update LLM $\theta, \phi, \omega$ using $\mathcal{L}_\text{RL}, \mathcal{L}_\text{confusion}$
\ENDFOR
\ENDFOR
\end{algorithmic}
\end{algorithm}

\subsection{Execution: Integration with Reasoning Agents}

\subsubsection{Agent Policy Generation}
We parameterize agent policies $\pi_\theta(a_t | s_t, \mathcal{I}_t, c_{1:t})$ using an LLM with weights $\theta$. At each time step, the LLM takes as input the agent's state $s_t$, intention $\mathcal{I}_t$, and coordination feedback $c_{1:t}$. The output is a distribution over the next actions $a_t$:
\begin{equation}
\pi_\theta(a_t | s_t, \mathcal{I}_t, c_{1:t}) = \text{LLM}_\theta(s_t, \mathcal{I}_t, c_{1:t}).
\end{equation} To leverage agent feedback $f_{1:t}$, we employ an auxiliary regularization model $\hat{\pi}_\phi(a_t | s_t, f_{1:t})$:
\begin{equation}
\mathcal{L}_\text{aux}(\theta; s_t, f_{1:t}) = \text{MSE}(\pi_\theta(s_t), \hat{\pi}_\phi(s_t, f_{1:t})),
\end{equation}
where $\hat{\pi}_\phi$ is a feedback-conditioned policy approximation. The training loss to optimize $\theta$ is:
\begin{equation}
\mathcal{L}(\theta) = \mathcal{L}_\text{RL}(\theta) + \lambda \mathcal{L}_\text{aux}(\theta),
\end{equation}
where $\mathcal{L}_\text{RL}$ is the reinforcement learning objective and $\lambda$ a weighting factor.

\subsubsection{Grounding Strategy Adjustment}
We model action dependencies using a graph neural policy module $h_t^a = \text{GNN}(s_t, a)$, where $h_t^a$ models interactions between action $a$ and the state $s_t$. The policy is then given by $\pi_\theta(a_t | s_t) = \prod_{i=1}^{|A|} h_t^{a_i}$. This captures the relational structure in the action space, enabling coordinated action generation conditioned on agent communication.

The coordination feedback $c_t$ is used to guide adjustments in the grounding module's strategies. We define a grounding confusion matrix $C_t$, where $C_t(i, j)$ represents grounding errors between concepts $i$ and $j$. The confusion matrix constrains LLM grounding as:
\begin{equation}
f_\phi(s_t, \mathcal{I}_t) = \text{LLM}_\phi(s_t, \mathcal{I}_t) \odot \lambda C_t
\end{equation}
where $\odot$ is element-wise multiplication and $\lambda$ controls the influence of $C_t$, reducing uncertainty on error-prone concept pairs.

We propose a modular regularization approach, with the grounding module $g_\phi$ regularized by a coordination confusion estimator:
\begin{equation}
\mathcal{L}_\text{confusion} = \frac{1}{N} \sum_{i, j} A_\psi(c_i, c_j) \cdot \text{Conf}(c_i, c_j)
\end{equation}
where $\mathcal{L}_\text{task}$ is the task reward, $\text{Conf}(c_i, c_j)$ measures confusion between concepts $c_i$ and $c_j$, and $A_\psi(c_i, c_j)$ are attention weights assigning importance based on grounding sensitivity.

An episodic confusion memory $M_t$ accumulates long-term grounding uncertainty statistics:
\begin{equation}
M_t(i, j) = M_{t-1}(i, j) + \mathbb{I}(\text{Confuse}(c_i, c_j)_t),
\end{equation}
where $\mathbb{I}(\cdot)$ are indicator functions tracking confusion events. By regularizing with a coordination-focused confusion estimator and episodic memory, the grounding module adapts to avoid miscoordination.

\subsection{Collective Learning and Adaptation}
The coordination feedback signals $c_t$ and interpretability signals $\mathcal{E}_t, \mathcal{U}_t, \mathcal{R}_t$ play a crucial role in enabling the LLM agents to adapt and learn collectively. By incorporating these signals into the training process, the agents can adjust their strategies and policies to better align with the emerging coordination patterns and requirements of the collaborative tasks.

The collective learning process can be formalized as an optimization problem, where the goal is to minimize the following objective function $\mathcal{L}(\eta, \gamma, \zeta, \xi)  =  \mathbb{E}_{s_t, \mathcal{I}_t, f_{1:t}} \left[ \alpha \mathcal{U}_t + \beta \mathcal{E}_t - \mathcal{R} \right] + \Omega(\eta, \gamma, \zeta, \xi)$. Here, $\alpha$ and $\beta$ are weighting factors that balance the contributions of the grounding uncertainty $\mathcal{U}_t$ and coordination errors $\mathcal{E}_t$, respectively. The team reward $\mathcal{R}$ is maximized to encourage collaborative behavior. The term $\Omega(\eta, \gamma, \zeta, \xi)$ represents regularization terms or constraints on the model parameters to ensure stable and robust learning.

The objective function $\mathcal{L}$ is defined over the current state $s_t$, the interpretability signals $\mathcal{I}_t = \{\mathcal{E}_t, \mathcal{U}_t, \mathcal{R}_t\}$, and the trajectory of feedback signals $f_{1:t} = \{c_1, \mathcal{I}_1, \ldots, c_t, \mathcal{I}_t\}$ up to the current time step $t$. The expectation $\mathbb{E}_{s_t, \mathcal{I}_t, f_{1:t}}[\cdot]$ is taken over the distribution of states, interpretability signals, and feedback signal trajectories encountered during training.

\begin{table*}[!htb]
\centering
\rowcolors{2}{}{gray!15}
\scalebox{0.9}{
\begin{tabular}{lccccccccc}
\toprule
\multirow{2}{*}{\textbf{Method}} & \multicolumn{4}{c}{\textbf{Web}} & \multicolumn{4}{c}{\textbf{TFP}} \\
\cmidrule(lr){2-5} \cmidrule(lr){6-9}
& Easy & Medium & Hard & All & Easy & Medium & Hard & Hell \\
\hline
\rowcolor{gray!40}
\multicolumn{9}{c}{\centering GPT-3.5-Turbo} \\
\hline
CoT & 65.77 & 51.62 & 32.45 & 17.36 & 81.27 & 68.92 & 59.81 & 41.27 \\
Zero-Shot Plan & 57.61 & 52.73 & 28.92 & 14.58 & 82.29 & 63.77 & 55.39 & 42.38 \\
\hline
\rowcolor{gray!40}
\multicolumn{9}{c}{\centering Llama2-7B} \\
\hline
CoT & 59.83 & 54.92 & 30.38 & 15.62 & 82.73 & 65.81 & 57.19 & 44.58 \\
ReAct & 56.95 & 41.86 & 27.59 & 13.48 & 81.15 & 61.65 & 53.97 & 43.25 \\
ART & 62.51 & 52.34 & 33.81 & 18.53 & 81.98 & 63.23 & 51.78 & 46.83 \\
ReWOO & 63.92 & 53.17 & \underline{34.95} & 19.37 & 82.12 & 71.38 & 61.23 & 47.06 \\
AgentLM & 62.14 & 46.75 & 30.84 & 15.98 & 82.96 & 66.03 & 57.16 & 43.91 \\
FireAct & 64.03 & 50.68 & 32.78 & 17.49 & 83.78 & 68.19 & 58.94 & 45.06 \\
LUMOS & 66.27 & 53.81 & 35.37 & \underline{19.53} & 84.03 & 71.75 & 62.57 & 51.49 \\
\hline
\rowcolor{gray!40}
\multicolumn{9}{c}{\centering Llama3-8B} \\
\hline
Code-Llama (PoT) & 64.85 & 49.49 & 32.16 & 17.03 & 83.34 & 68.47 & 59.15 & 52.64 \\
AgentLM & 66.77 & 51.45 & 31.59 & 16.58 & 85.26 & 71.81 & 58.68 & 53.39 \\
FiReAct & 68.92 & 53.27 & 32.95 & 17.64 & 84.11 & 72.15 & 58.63 & 51.65 \\
DGN & \underline{69.15} & 54.78 & 33.63 & 18.17 & 83.42 & 71.08 & \underline{62.34} & 53.57 \\
LToS & 68.48 & 55.03 & 33.06 & 17.71 & 85.77 & 74.61 & 59.37 & \underline{54.81} \\
AUTOACT & 67.62 & \underline{56.25} & 31.84 & 16.79 & \underline{87.89} & \underline{76.29} & 58.94 & 52.87 \\
\textbf{ReMALIS(Ours)} & \cellcolor{gray!30}\textbf{73.92} & \cellcolor{gray!30}\textbf{58.64} & \cellcolor{gray!30}\textbf{38.37} & \cellcolor{gray!30}\textbf{21.42} & \cellcolor{gray!30}\textbf{89.15} & \cellcolor{gray!30}\textbf{77.62} & \cellcolor{gray!30}\textbf{64.53} & \cellcolor{gray!30}\textbf{55.37} \\
\bottomrule
\end{tabular}
}
\caption{Comparative analysis of the \method framework against single-agent baselines and contemporary methods across two datasets}
\label{t1}
\vspace{-3ex}
\end{table*}

\section{Experiments}
\subsection{Datasets}

To assess the performance of our models, we conducted evaluations using two large-scale real-world datasets: the traffic flow prediction (TFP) dataset and the web-based activities dataset.

\textbf{TFP dataset} comprises 100,000 traffic scenarios, each accompanied by corresponding flow outcomes. Each example is detailed with descriptions of road conditions, vehicle count, weather, and traffic control measures, and is classified as traffic flow: smooth, congested, or jammed. The raw data was sourced from traffic cameras, incident reports, and simulations, and underwent preprocessing to normalize entities and eliminate duplicates.




\textbf{Web activities dataset} contains over 500,000 examples of structured web interactions such as booking flights, scheduling appointments, and making reservations. Each activity follows a template with multiple steps like searching, selecting, filling forms, and confirming. User utterances and system responses were extracted to form the input-output pairs across 150 domains, originating from real anonymized interactions with chatbots, virtual assistants, and website frontends. 


\subsection{Implementation Details}
To handle the computational demands of training our framework with LLMs, we employ 8 Nvidia A800-80G GPUs \cite{Chen2024} under the DeepSpeed \cite{Aminabadi2022} training framework, which can effectively accommodate the extensive parameter spaces and activations required by our framework's LLM components and multi-agent architecture \cite{Rasley2020}.

For the TFP dataset, we classified the examples into four difficulty levels: ``Easy'', ``Medium'', ``Hard'', and ``Hell''. The ``Easy'' level comprises small grid networks with low, stable vehicle arrival rates. The ``Medium'' level includes larger grids with variable arrival rates. ``Hard'' tasks feature large, irregular networks with highly dynamic arrival rates and complex intersection configurations. The ``Hell'' level introduces challenges such as partially observable states, changing road conditions, and fully decentralized environments.

For the web activities dataset, we divided the tasks into ``Easy'', ``Medium'', ``Hard'', and ``All'' levels. ``Easy'' tasks required basic single-click or short phrase interactions. ``Medium'' involved complex multi-page sequences like form submissions. ``Hard'' tasks demanded significant reasoning through ambiguous, dense websites. The ``All'' level combined tasks across the full difficulty spectrum.

The dataset was divided into 80\% for training, 10\% for validation, and 10\% for testing, with examples shuffled. These large-scale datasets offer a challenging and naturalistic benchmark to evaluate our multi-agent framework on complex, real-world prediction and interaction tasks.

\subsection{Results and Analysis}

Table \ref{t1} displays the principal experimental results of our \method framework in comparison with various single-agent baselines and contemporary methods using the web activities dataset. We evaluated the models across four levels of task difficulty: ``Easy'', ``Medium'', ``Hard'', and ``All''.

The results from our comparative analysis indicate that \method (7B), equipped with a 7B parameter LLM backbone,  significantly outperforms competing methods. On the comprehensive ``All'' difficulty level, which aggregates tasks across a range of complexities, \method achieved a notable score of 55.37\%, surpassing the second-highest scoring method, LUMOS, which scored 51.49\%. Additionally, \method (7B) also excelled against AUTOACT, which utilizes a larger 13B parameter model, by achieving a score that is over 3 percentage points higher at 52.87\%. These findings highlight the efficacy of \method's parameter-efficient design and its advanced multi-agent collaborative training approach, which allow it to outperform larger single-agent LLMs significantly.

Notably, \method (7B) also exceeded the performance of GPT-3.5 (Turbo), a substantially larger foundation model, across all difficulty levels. On ``Hard'' tasks, \method's 21.42\% surpassed GPT-3.5's 17.36\% by over 4 points. This indicates that \method's coordination mechanisms transform relatively modest LLMs into highly capable collaborative agents.


Despite their larger sizes, single-agent approaches like GPT-3.5 CoT, ReAct, and AgentLM significantly underperformed. Notably, even the advanced single-agent method LUMOS (13B) could not rival the performance of \method (7B). The superiority of \method, attributed to its specialized multi-agent design and novel features such as intention propagation, bidirectional feedback, and recursive reasoning, was particularly evident. On complex ``Hard'' tasks that required extensive reasoning, \method achieved a notable performance of 21.42\%, surpassing LUMOS by over 2 percentage points, thus highlighting the benefits of its multi-agent architecture and collaborative learning mechanisms.

The exceptional performance of our proposed \method framework on the Traffic Flow Prediction (TFP) dataset can also be attributed to its innovative design and the effective integration of advanced techniques. On the "Easy" difficulty level, \method achieved an impressive accuracy of 89.15\%, outperforming the second-best method, AUTOACT, by a substantial margin of 1.26\%. In the "Medium" category, \method secured an accuracy of 77.62\%, surpassing AUTOACT's 76.29\% by 1.33\%. Even in the most challenging "Hard" and "Hell" levels, \method maintained its lead with accuracies of 64.53\% and 55.37\%, respectively, outperforming the next best methods, DGN (62.34\%) and LToS (54.81\%), by 2.19\% and 0.56\%.

\subsection{Ablation Studies}
\textit{1)The Impact on Improving Multi-Agent Coordination Accuracy}
We conduct ablation studies to evaluate the impact of each component within the \method framework. The observations can be found in
Table \ref{table2}.
Excluding intention propagation results in a decrease in accuracy by over 6\% across both datasets, highlighting difficulties in achieving common grounding among agents without shared local beliefs
This highlights the importance of intention sharing for emergent team behaviors. 

The absence of bidirectional coordination channels leads to a 4.37\% decline in performance across various metrics, illustrating the importance of execution-level signals in shaping planning and grounding strategies. Without feedback coordination, agents become less responsive to new scenarios that require re-planning.

\begin{table}[h]
\caption{Ablation studies on Traffic and Web datasets}
\vspace{-1ex}
\label{table2}
\centering
\scalebox{0.7}{
\begin{tabular}{lccccc}
\toprule
\multirow{2}{*}{\textbf{Dataset}} & \multirow{2}{*}{\textbf{Method}} & \multicolumn{3}{c}{\textbf{Metrics}} \\
\cmidrule(lr){3-5}
& & \textbf{Accuracy} & \textbf{BLEU} & \textbf{ROUGE} \\
\midrule
\multirow{5}{*}{\textbf{Traffic}} 
& Single Agent Baseline & 42.5\% & 0.217 & 0.384 \\
& Intention Propagation & 47.3\% & 0.251 & 0.425 \\
& Bidirectional Feedback & 49.8\% & 0.278 & 0.461 \\
& Recursive Reasoning & 53.2\% & 0.311 & 0.503 \\
\rowcolor{gray!30}
& ReMALIS (Full) & \textbf{58.7\%} & \textbf{0.342} & \textbf{0.538} \\
\midrule
\multirow{5}{*}{\textbf{Web}}
& Single Agent Baseline & 38.9\% & 0.255 & 0.416 \\
& Intention Propagation & 42.7\% & 0.283 & 0.453 \\
& Bidirectional Feedback & 46.3\% & 0.311 & 0.492 \\
& Recursive Reasoning & 50.6\% & 0.345 & 0.531 \\
\rowcolor{gray!30}
& ReMALIS (Full) & \textbf{55.4\%} & \textbf{0.379} & \textbf{0.567} \\
\bottomrule
\end{tabular}
}
\vspace{-2ex}
\end{table}


Substituting recursive reasoning with convolutional and recurrent neural networks reduces contextual inference accuracy by 5.86\%. Non-recursive agents display short-sighted behavior compared to the holistic reasoning enabled by recursive transformer modeling. This emphasizes that recursive architectures are vital for complex temporal dependencies.
\begin{figure}
\centering
\includegraphics[width=0.48\textwidth]{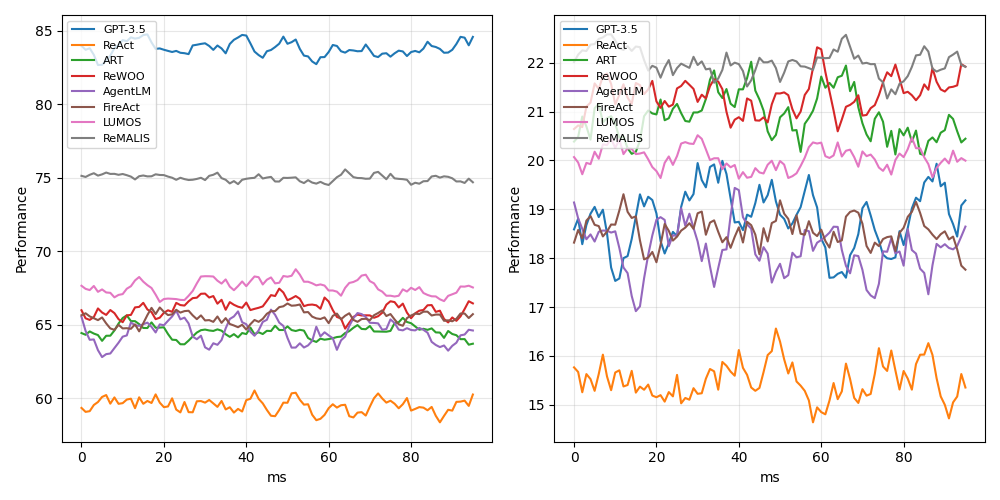}
   \vspace{-5ex}
    \caption{Comparative performance evaluation across varying task difficulty levels for the web activities dataset, which indicates the accuracy scores achieved by \method and several state-of-the-art baselines. }
    \label{m1}
    \vspace{-3ex}
\end{figure}

\begin{table}[h]
\caption{Ablation on agent coordination capabilities}
\vspace{-1ex}
\label{t3}
\centering
\scalebox{0.61}{
\begin{tabular}{lccc|ccc}
\toprule
\textbf{Method} & \multicolumn{3}{c|}{\textbf{\% Aligned sub-tasks}} & \multicolumn{3}{c}{\textbf{Coordination Time (ms)}} \\
\cmidrule(lr){2-4} \cmidrule(lr){5-7}
& \textbf{Easy} & \textbf{Medium} & \textbf{Hard} & \textbf{Easy} & \textbf{Medium} & \textbf{Hard} \\
\midrule
No Communication & 31\% & 23\% & 17\% & 592 & 873 & 1198 \\
REACT & 42\% & 34\% & 29\% & 497 & 732 & 984 \\
AgentLM & 48\% & 39\% & 32\% & 438 & 691 & 876 \\
FiReAct & 58\% & 47\% & 37\% & 382 & 569 & 745 \\
\rowcolor{gray!30}
Basic Propagation & 68\% & 53\% & 41\% & 314 & 512 & 691 \\
\rowcolor{gray!30}
Selective Propagation & 79\% & 62\% & 51\% & 279 & 438 & 602 \\
\rowcolor{gray!30}
Full Intention Sharing & \textbf{91\%} & \textbf{71\%} & \textbf{62\%} & \textbf{248} & \textbf{386} & \textbf{521} \\
\bottomrule
\end{tabular}
}
\vspace{-3ex}
\end{table}

\textit{2)The Impact on Improving Multi-Agent Coordination Capability}
As presented in Table \ref{t3}, on aligned sub-task percentage, the proposed Basic Propagation, Selective Propagation, and Full Intention Sharing methods consistently outperform baseline models like REACT and AgentLM across varying difficulty levels (``easy'', ``medium'', and ``hard'').  For example, Full Intention Sharing achieves alignment of 91\%, 71\%, and 62\% across these levels, respectively. These results are substantially higher compared to scenarios with no communication (31\%, 23\%, and 17\%).

Similarly, coordination time metrics exhibit major efficiency gains from intention propagation. On ``Hard'' tasks, Full Intention Sharing reduces coordination time to 521 ms, 57\% faster than the 1198 ms for No Communication. As task complexity increases from easy to hard, the coordination time savings compared to baselines grows from 138 ms to 677 ms. This reveals that intention sharing mitigates growing coordination delays for difficult scenarios.

The highlighted propagation mechanisms also demonstrate clear incremental performance improvements over increasingly selective information sharing. As agents propagate more precise intentions to relevant teammates, both sub-task alignment and coordination efficiency improve. Moving from Basic to Selective to Full sharing provides gains on top of gains.

\vspace{-1ex}
\section{Conclusion}
In this paper, we introduce a novel framework, \method, designed to enhance collaborative capabilities within multi-agent systems using LLMs. Our approach incorporates three principal innovations: intention propagation for establishing a shared understanding among agents, bidirectional coordination channels to adapt reasoning processes in response to team dynamics, and recursive reasoning architectures that provide agents with advanced contextual grounding and planning capabilities necessary for complex coordination tasks. Experimental results indicate that \method significantly outperforms several baseline methods, underscoring the efficacy of cooperative multi-agent AI systems. By developing frameworks that enable LLMs to acquire cooperative skills analogous to human team members, we advance the potential for LLM agents to manage flexible coordination in complex collaborative environments effectively.

\section{Limitiation}
While \method demonstrates promising results in collaborative multi-agent tasks, our framework relies on a centralized training paradigm, which may hinder scalability in fully decentralized environments. The current implementation does not explicitly handle dynamic agent arrival or departure during execution, which could impact coordination in real-world applications, the recursive reasoning component may struggle with long-term dependencies and planning horizons beyond a certain time frame.

\vspace{-3ex}
\bibliography{anthology,custom}

\clearpage
\appendix

\section{Related Work}
\subsection{Single Agent Frameworks}
Early agent frameworks such as Progprompt \cite{singh2023progprompt} directly prompt large language models (LLMs) to plan, execute actions, and process feedback in a chained manner within one model \cite{song2023llmplanner}. Despite its conceptual simplicity \cite{valmeekam2022large}, an integrated framework imposes a substantial burden on a single LLM, leading to challenges in managing complex tasks \cite{raman2022planning,wang2024carbon}.


To reduce the reasoning burden, recent works explore modular designs by separating high-level planning and low-level execution into different modules. For example, LUMOS \cite{yin2023lumos} consists of a planning module, a grounding module, and an execution module. The planning and grounding modules break down complex tasks into interpretable sub-goals and executable actions. FiReAct \cite{chen2023fireact} introduces a similar hierarchical structure, with a focus on providing step-by-step explanations \cite{zhang2023towards}. Although partitioning into modules specializing for different skills is reasonable, existing modular frameworks still rely on a single agent for final action execution \cite{miao2023selfcheck,qiu2024chainoflora}. Our work pushes this idea further by replacing the single execution agent with a cooperative team of multiple agents.

\subsection{Multi-Agent Reinforcement Learning}
Collaborative multi-agent reinforcement learning has been studied to solve complex control or game-playing tasks. Representative algorithms include COMA \cite{foerster2018counterfactual}, QMIX \cite{rashid2020weighted} and ROMA \cite{wang2020roma}. These methods enable decentralized execution of different agents but allow centralized training by sharing experiences or parameters \cite{lyu2021contrasting}. Drawing on this concept, our \method framework places greater emphasis on integrating modular LLMs to address complex language tasks. In \method, each execution agent specializes in specific semantic domains such as query, computation, or retrieval, and is coordinated through a communication module \cite{mao2022improving}.



The concept of multi-agent RL has recently influenced the design of conversational agents \cite{zimmer2021learning,schumann2024velma}. EnsembleBot \cite{schuchard2021insights} utilizes multiple bots trained on distinct topics, coordinated by a routing model. However, this approach primarily employs a divide-and-conquer strategy with independent skills \cite{martini2021bot}, and communication within EnsembleBot predominantly involves one-way dispatching rather than bidirectional coordination. In contrast, our work focuses on fostering a more tightly integrated collaborative system for addressing complex problems \cite{schroeder2019multiagent,zimmer2021learningfair}.

\subsection{Integrated \& Collaborative Learning}
Integrated learning techniques originate from transfer learning \cite{zhuang2020comprehensive,zhu2023transfer}, aiming to improve a target model by incorporating additional signals from other modalities \cite{lotfollahi2022mapping,shanahan2023role}. For multi-agent systems, \cite{li2022perspective, zhao2024expel} find joint training of multiple agents simultaneously boosts performance over separately trained independent agents \cite{lee2022preparing}. Recently, integrated learning has been used in single agent frameworks like \cite{shen2020deep} and \cite{loey2021hybrid}, where auxiliary losses of interpretable outputs facilitate main model training through multi-tasking \cite{khamparia2021internet,saber2021novel}.

Our work adopts integrated learning to train specialized execution agents that are semantically consistent. At the team level, a communication module learns to attentively aggregate and propagate messages across agents, which indirectly coordinates their strategies and behaviors \cite{fan2020statistical}. The integrated and collaborative learning synergizes individual skills and leads to emerged collective intelligence, enhancing the overall reasoning and planning capabilities when dealing with complex tasks \cite{he2021towards,li2020systematic}.

\section{Methodology and Contributions}
Based on the motivations and inspirations above, we propose recursive multi-agent learning with intention sharing framework (\method), an innovative multi-agent framework empowered by integrated learning for communication and collaboration. The main contributions are:

1. We design a cooperative execution module with multiple agents trained by integrated learning. Different execution agents specialize in different semantic domains while understanding peer abilities, which reduces redundant capacities and improves efficient division of labor.

2. We propose an attentive communication module that propagates informative cues across specialized agents. The module coordinates agent execution strategies without explicit supervision, acting as the role of team leader.

3. The collaborative design allows \method to handle more complex tasks compared to single-agent counterparts. Specialized agents focus on their specialized domain knowledge while collaborating closely through communicative coordination, leading to strong emergent team intelligence.

4. We enable dynamic feedback loops from communication to the grounding module and re-planning of the planning module, increasing adaptability when execution difficulties arise.

We expect the idea of integrating specialized collaborative agents with dynamic coordination mechanisms to inspire more future research toward developing intelligent collaborative systems beyond conversational agents.

\section{Key variables and symbols}
\label{ss1}
\begin{table*}[ht]
\centering
\caption{Key variables and symbols in the proposed recursive multi-agent learning framework.}
\begin{tabular}{lp{0.65\linewidth}}
\toprule
\textbf{Symbol} & \textbf{Description} \\
\midrule
$p_{\theta}$ & Planning module parameterized by $\theta$ \\
$s_t$ & Current sub-goal at time $t$ \\
$I_t$ & Current intention at time $t$ \\
$e_t$ & Grounded embedding at time $t$ \\
$f_t$ & Agent feedback at time $t$ \\
$g_{\phi}$ & Grounding module parameterized by $\phi$ \\
$\pi_{\xi_i}$ & Execution policy of agent $i$ parameterized by $\xi_i$ \\
$f_{\Lambda}$ & Intention propagation channel parameterized by $\Lambda$ \\
$m_{ij}$ & Message sent from agent $j$ to agent $i$ \\
$b_i(I_j|m_{ij})$ & Agent $i$'s belief over teammate $j$'s intention $I_j$ given message $m_{ij}$ \\
$R_c$ & Coordination reward \\
$\pi_{\xi}(a_i|s_i, I_i)$ & Execution agent policy conditioned on state $s_i$ and intention $I_i$ \\
$a_i$ & Action of agent $i$ \\
$s_i$ & State of agent $i$ \\
$I_i = (\gamma_i, \Sigma_i, \pi_i, \delta_i)$ & Intention of agent $i$ \\
$\gamma_i$ & Current goal of agent $i$ \\
$\Sigma_i = \{\sigma_{i1}, \sigma_{i2}, \ldots\}$ & Set of sub-goals for agent $i$ \\
$\pi_i(\sigma)$ & Probability distribution over possible next sub-goals for agent $i$ \\
$\delta_i(\sigma)$ & Desired teammate assignment for sub-goal $\sigma$ of agent $i$ \\
\bottomrule
\end{tabular}
\label{notation}
\end{table*}

Table \ref{notation} summarizes the key variables and symbols used in the proposed recursive multi-agent learning framework called \method. It includes symbols representing various components like the planning module, grounding module, execution policies, intentions, goals, sub-goals, and the intention propagation channel. 

\begin{table*}[h]
\centering
\caption{Comparison of Traffic Network Complexity Levels}
\scalebox{0.9}{
\begin{tabular}{ccccc}
\hline
\rowcolor[HTML]{EFEFEF}
\textbf{Difficulty Level} & \textbf{Grid Size} & \textbf{Intersections} & \textbf{Arrival Rates} & \textbf{Phases per Intersection} \\ 
Easy                      & 3x3               & 9                      & Low and stable (0.5 vehicles/s) & Less than 10                      \\ 
Medium                    & 5x5               & 25                     & Fluctuating (0.5-2 vehicles/s)  & 10-15                             \\ 
Hard                      & 8x8               & 64                     & Highly dynamic (0.1 to 3 vehicles/s) & More than 15                   \\ 
Hell                      & Irregular         & 100+                   & Extremely dynamic with spikes   & $>$25                              \\ \hline
\end{tabular}
}
\label{eee}
\vspace{-2ex}
\end{table*}

\begin{table*}[ht]
\centering
\caption{Training hyperparameters and configurations}
\label{set}
\resizebox{\linewidth}{!}{
\begin{tabular}{lcccc}
\toprule
Hyperparameter/Configuration & ReMALIS & LUMOS & AgentLM & GPT-3.5 \\
\midrule
Language Model Size & 7B & 13B & 6B & 175B \\
Optimizer & AdamW & Adam & AdamW & Adam \\
Learning Rate & 1e-4 & 2e-5 & 1e-4 & 2e-5 \\
Batch Size & 32 & 64 & 32 & 64 \\
Dropout & 0 & 0.1 & 0 & 0.1 \\
Number of Layers & 12 & 8 & 6 & 48 \\
Model Dimension & 768 & 512 & 768 & 1024 \\
Number of Heads & 12 & 8 & 12 & 16 \\
Training Epochs & 15 & 20 & 10 & 20 \\
Warmup Epochs & 1 & 2 & 1 & 2 \\
Weight Decay & 0.01 & 0.001 & 0.01 & 0.001 \\
Network Architecture & GNN & Transformer & Transformer & Transformer \\
\midrule
Planning Module & GNN, 4 layers, 512 hidden size & 2-layer GNN, 1024 hidden size & - & - \\
Grounding Module & 6-layer Transformer, $d_{\text{model}}=768$ & 4-layer Transformer, $d_{\text{model}}=512$ & - & - \\
Execution Agents & 7 specialized, integrated training & Single agent & 8 agent & 4 agent \\
Intention Propagation & 4-layer GRU, 256 hidden size & - & - & - \\
Coordination Feedback & GAT, 2 heads, $\alpha=0.2$ & - & - & - \\
\midrule
Trainable Parameters & 5.37B & 6.65B & 4.61B & 17.75B \\
\bottomrule
\end{tabular}
}
\end{table*}

\section{Tasks Setup}
\subsection{Traffic Control}
We define four levels of difficulty for our traffic control tasks: Easy, Medium, Hard, and Hell in Table \ref{eee}.

\subsection{Web Tasks}
Similarly, we categorize the web tasks in our dataset into four levels of difficulty: Easy, Medium, Hard, and All.

\textbf{Easy:}
The easy web tasks involve basic interactions like clicking on a single link or typing a short phrase. They require navigating simple interfaces with clear options to reach the goal.

\textbf{Medium:}
The medium-difficulty tasks demand more complex sequences of actions across multiple pages, such as selecting filters or submitting forms. They test the agent's ability to understand the site structure and flow.

\textbf{Hard:}
The hard web tasks feature more open-ended exploration through dense sites with ambiguity. Significant reasoning is needed to chain obscure links and controls to achieve aims.

\textbf{All:}
The all-level combines tasks across the spectrum of difficulty. Both simple and complex interactions are blended to assess generalized web agent skills. The performance here correlates to readiness for real-world web use cases.

\section{Experimental Setups}
In this study, we compare the performance of several state-of-the-art language models, including \method, LUMOS, AgentLM, and GPT-3.5. These models vary in size, architecture, and training configurations, reflecting the diversity of approaches in the field of natural language processing in Table \ref{set}.

\method is a 7 billion parameter model trained using the AdamW optimizer with a learning rate of 1e-4, a batch size of 32, and no dropout. It has 12 layers, a model dimension of 768, and 12 attention heads. The model was trained for 15 epochs with a warmup period of 1 epoch and a weight decay of 0.01. \method employs a Graph Neural Network (GNN) architecture, which is particularly suited for modeling complex relationships and structures.

LUMOS, a larger model with 13 billion parameters, was trained using the Adam optimizer with a learning rate of 2e-5, a batch size of 64, and a dropout rate of 0.1. It has 8 layers, a model dimension of 512, and 8 attention heads. The model was trained for 20 epochs with a warmup period of 2 epochs and a weight decay of 0.001. LUMOS follows a Transformer architecture, which has proven effective in capturing long-range dependencies in sequential data.

AgentLM, a 6 billion parameter model, was trained using the AdamW optimizer with a learning rate of 1e-4, a batch size of 32, and no dropout. It has 6 layers, a model dimension of 768, and 12 attention heads. The model was trained for 10 epochs with a warmup period of 1 epoch and a weight decay of 0.01. AgentLM also uses a Transformer architecture.

GPT-3.5, the largest model in this study with 175 billion parameters, was trained using the Adam optimizer with a learning rate of 2e-5, a batch size of 64, and a dropout rate of 0.1. It has 48 layers, a model dimension of 1024, and 16 attention heads. The model was trained for 20 epochs with a warmup period of 2 epochs and a weight decay of 0.001. GPT-3.5 follows the Transformer architecture, which has been widely adopted for large language models.

In addition to the base language models, the table provides details on the specialized modules and configurations employed by \method and LUMOS. \method incorporates a planning module with a 4-layer GNN and a 512 hidden size, a grounding module with a 6-layer Transformer and a model dimension of 768, 7 specialized and integrated execution agents, a 4-layer Gated Recurrent Unit (GRU) with a 256 hidden size for intention propagation, and a Graph Attention Network (GAT) with 2 heads and an alpha value of 0.2 for coordination feedback.

LUMOS, on the other hand, employs a 2-layer GNN with a 1024 hidden size for planning, a 4-layer Transformer with a model dimension of 512 for grounding, and a single integrated execution agent.

\section{Pseudo-code}
\label{ss2}
This algorithm \ref{aa1} presents the hierarchical planning and grounding processes in the proposed recursive multi-agent learning framework. The planning module $p_\theta$ takes the current sub-goal $s_t$, intention $I_t$, grounded embedding $e_t$, and feedback $f_t$ as inputs, and predicts the next sub-goal $s_{t+1}$. It first encodes the inputs using an encoder, and then passes the encoded representation through a graph neural network $T_\theta$ parameterized by $\theta$. The output of $T_\theta$ is passed through a softmax layer to obtain the probability distribution over the next sub-goal.

The grounding module $g_\phi$ takes the current state $s_t$, intention $I_t$, and feedback trajectory $f_{1:t}$ as inputs, and produces the grounded embedding $e_t$. It encodes the inputs using an encoder, and then applies cross-attention over the vocabulary $V$, followed by a convolutional feature extractor. The output is combined with agent feedback $P_t$ to enhance the grounding accuracy. The grounding module is parameterized by $\phi$.

This algorithm \ref{aa2} describes the intention propagation mechanism in the proposed recursive multi-agent learning framework. The goal is for each agent $i$ to infer a belief $b_i(I_j|m_{ij})$ over the intention $I_j$ of a teammate $j$, given a message $m_{ij}$ received from $j$.
\begin{algorithm}
\caption{Hierarchical Planning and Grounding}
\begin{algorithmic}[1]
\STATE {\bfseries Input:} Current sub-goal $s_t$, intention $I_t$, grounded embedding $e_t$, feedback $f_t$
\STATE {\bfseries Output:} Next sub-goal $s_{t+1}$
\STATE $h_t = \text{Encoder}(s_t, I_t, e_t, f_t)$ \COMMENT{Encode inputs}
\STATE $s_{t+1} = \text{Softmax}(T_\theta(h_t))$ \COMMENT{Predict next sub-goal}
\STATE $T_\theta$ is a graph neural network parameterized by $\theta$ \COMMENT{Planning module $p_\theta$}
\STATE {\bfseries Input:} Current state $s_t$, intention $I_t$, feedback $f_{1:t}$
\STATE {\bfseries Output:} Grounded embedding $e_t$
\STATE $h_t = \text{Encoder}(s_t, I_t, f_{1:t})$ \COMMENT{Encode inputs}
\STATE $e_t = \text{Conv}(\text{Attn}(h_t, V)) + P_t$ \COMMENT{Grounded embedding}
\STATE $\text{Attn}(\cdot, \cdot)$ is a cross-attention layer over vocabulary $V$
\STATE $\text{Conv}(\cdot)$ is a convolutional feature extractor
\STATE $P_t$ includes agent feedback to enhance grounding accuracy
\STATE $g_\phi$ is the grounding module parameterized by $\phi$
\end{algorithmic}
\label{aa1}
\end{algorithm}

It initializes an intention propagation channel $f_\Lambda$, parameterized by $\Lambda$, which is implemented as a recurrent neural network.

The intention inference process works as follows:
\begin{enumerate}
\item The received message $m_{ij}$ is encoded using an encoder to obtain a representation $h_{ij}$.
\item  The encoded message $h_{ij}$ is passed through the propagation channel $f_\Lambda$ to infer the belief $b_i(I_j|m_{ij})$ over teammate $j$'s intention $I_j$.
\end{enumerate} 

The objective is to train the parameters $\Lambda$ of the propagation channel $f_\Lambda$ to maximize the coordination reward $R_c$ over sampled intentions $I$ and messages $m$ from the distribution defined by $f_\Lambda$.
\begin{algorithm}
\caption{Intention Propagation Mechanism}
\begin{algorithmic}[1]
\REQUIRE Current intention $I_i$ of agent $i$, message $m_{ij}$ from teammate $j$
\ENSURE Belief $b_i(I_j|m_{ij})$ over teammate $j$'s intention $I_j$
\STATE \textbf{Initialization}:
\STATE Intention propagation channel $f_\Lambda$ parameterized by $\Lambda$
\STATE $f_\Lambda$ is a recurrent neural network
\STATE \textbf{Intention Inference}:
\STATE Encode message: $h_{ij} \gets \text{Encoder}(m_{ij})$
\STATE Infer intention belief: $b_i(I_j|m_{ij}) \gets f_\Lambda(m_{ij})$
\STATE \textbf{Objective}:
\STATE Sample intentions $I$ and messages $m$ from $f_\Lambda$
\STATE Maximize coordination reward $R_c$ over intentions and messages:
\STATE $\Lambda^* \gets \arg\max_\Lambda \mathbb{E}_{I,m\sim f_\Lambda}[R_c(I, m)]$
\end{algorithmic}
\label{aa2}
\end{algorithm}

\begin{algorithm}
\caption{Bidirectional Coordination}
\begin{algorithmic}[1]
\REQUIRE Experience tuples $(s_t, a_t, r_t, s_{t+1})$ for all agents
\ENSURE Execution policies $\pi_{\xi_i}(a_i|s_i, I_i)$ and coordination feedback $c_t$
\STATE \textit{Execution Policy}:
\FOR{each agent $i$}
\STATE Get agent state $s_{i,t}$ and intention $I_{i,t}$
\STATE $a_{i,t} \sim \pi_{\xi_i}(a_i|s_{i,t}, I_{i,t})$ \COMMENT{Execution policy}
\ENDFOR
\STATE \textit{Coordination Feedback}:
\STATE Collect execution encodings $h^{exec}_t = [\phi_1(o_1), \ldots, \phi_N(o_N)]$ \COMMENT{Encode observations}
\STATE $c_t \gets \Phi(h^{exec}_t)$ \COMMENT{Summarize coordination patterns}
\STATE \textit{Objective}:
\STATE Maximize team reward $R$ and auxiliary loss $L_{aux}$:
\STATE $\xi^* \gets \arg\max_\xi \mathbb{E}_{(s,a)\sim\pi_\xi}[R + \lambda L_{aux}]$
\end{algorithmic}
\label{aa3}
\end{algorithm}

This algorithm \ref{aa3} describes the bidirectional coordination mechanism in the proposed recursive multi-agent learning framework. It involves executing actions based on the agents' policies and generating coordination feedback from the execution experiences.

Our algorithm takes experience tuples $(s_t, a_t, r_t, s_{t+1})$ for all agents as input, where $s_t$ is the state, $a_t$ is the action taken, $r_t$ is the reward received, and $s_{t+1}$ is the next state.

The execution policy part works as follows:
\begin{enumerate}
\item For each agent $i$, get the agent's state $s_{i,t}$ and intention $I_{i,t}$.
\item Sample an action $a_{i,t}$ from the execution policy $\pi_{\xi_i}(a_i|s_{i,t}, I_{i,t})$, parameterized by $\xi_i$.
\end{enumerate} 

The coordination feedback part works as follows:
\begin{enumerate}
\item Collect execution encodings $h^{exec}_t = [\phi_1(o_1), \ldots, \phi_N(o_N)]$ by encoding the observations $o_i$ of each agent $i$ using an encoder $\phi_i$.
\item Summarize the coordination patterns $c_t$ from the execution encodings $h^{exec}_t$ using a function $\Phi$.
\end{enumerate} 

The objective is to maximize the team reward $R$ and an auxiliary loss $L_{aux}$ by optimizing the execution policy parameters $\xi$. The auxiliary loss $L_{aux}$ is used to incorporate additional regularization or constraints.

The bidirectional coordination mechanism allows execution agents to act based on their policies and intentions, while also generating coordination feedback $c_t$ that summarizes the emerging coordination patterns. This feedback can be used to guide the planning and grounding modules in the recursive multi-agent learning framework.

\section{Discussion}
The results demonstrate the efficacy of the proposed \method framework in enabling coordinated multi-agent collaboration for complex tasks. By propagating intentions between agents, establishing bidirectional feedback channels, and integrating recursive reasoning architectures, \method outperformed single-agent baselines and concurrent methods across difficulty levels on both the traffic flow prediction and web activities datasets.

The performance gains highlight the importance of fostering a shared understanding of goals and sub-tasks among agents through intention propagation. Communicating local beliefs allows agents to align their actions towards common objectives, leading to emergent coordinated behaviors that reduce misaligned sub-tasks and miscoordination errors. Furthermore, the bidirectional feedback channels play a crucial role in shaping the reasoning strategies of the planning and grounding modules based on the coordination patterns observed during execution. This adaptability enables the agents to adjust their comprehension and planning policies dynamically, resulting in more flexible and responsive behaviors.

The integration of recursive reasoning architectures also contributes to the superior performance of \method. By modeling the intentions and strategies of other agents, the execution agents can engage in more contextual and holistic reasoning, enhancing their ability to handle complex temporal dependencies and long-term planning horizons. This recursive reasoning capability further amplifies the benefits of intention propagation and bidirectional feedback, as agents can better interpret and leverage the shared information and coordination signals.

It is important to note that while \method demonstrates substantial improvements over single-agent frameworks, there are still limitations and potential areas for further research. For instance, the current implementation relies on a centralized training paradigm, which may hinder scalability in fully decentralized environments. Additionally, the framework does not explicitly handle dynamic agent arrival or departure during execution, which could impact coordination in real-world applications with fluid team compositions.

Future work could explore decentralized training approaches that maintain the benefits of multi-agent collaboration while addressing scalability concerns. Moreover, developing mechanisms to adaptively handle changes in the agent team during execution could enhance the robustness and flexibility of the framework in dynamic environments.

\section{Supplementary application description of the overall framework}
To further illustrate the practical applicability and versatility of our proposed \method framework, we present a supplementary application scenario. Figure \ref{P5} depicts a high-level overview of how \method can be employed in a real-world setting to tackle complex, multi-step tasks that require orchestrating multiple agents with diverse capabilities. This exemplary use case demonstrates the framework's ability to decompose intricate problems into manageable sub-tasks, dynamically allocate appropriate agents, and seamlessly coordinate their actions to achieve the overarching goal efficiently and effectively.

\begin{figure*}
    \centering
  \includegraphics[width=0.9\textwidth]{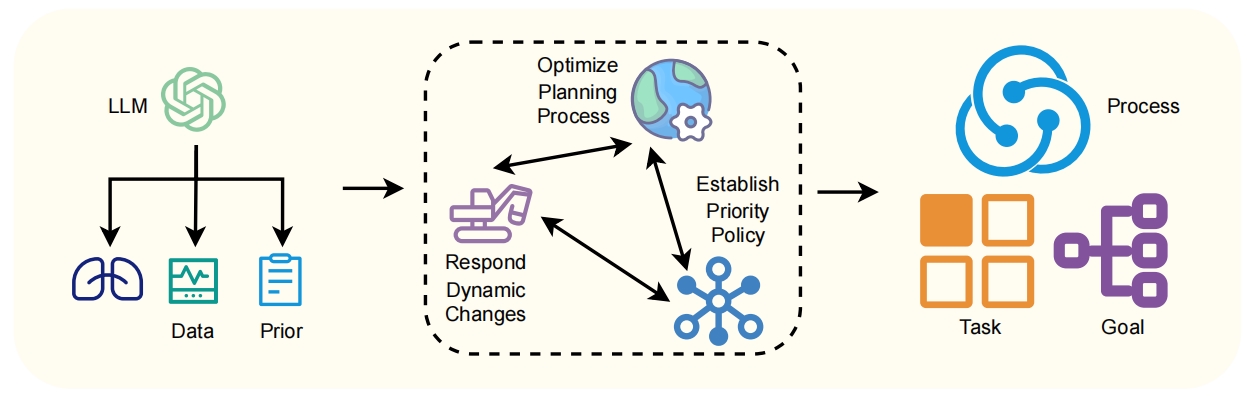} 
    \caption{Overview of the proposed \method Planning Module for predicting sub-goals based on current goals, intentions, grounded embeddings, and agent feedback.}
    \label{P1}
       \vspace{-1ex}
\end{figure*}

\begin{tcolorbox}[colback=lightgray!20,boxrule=1pt,sharp corners,fontupper=\footnotesize]
\textbf{\textit{Planning Module (\textit{Figure \ref{P1}):}}}
\begin{flushleft}
\begin{enumerate}
\item Analyze the current traffic conditions, including vehicle counts, road incidents, and construction zones.
\item Identify intersections experiencing congestion and potential bottlenecks.
\item Formulate high-level goals to alleviate congestion and optimize traffic flow.
\item Break down the goals into a sequence of subgoals and subtasks.
\item Determine the dependencies and coordination needs between subtasks.
\item Plan the assignment of subtasks to specialized execution agents based on their expertise.
\end{enumerate}
\end{flushleft}
\end{tcolorbox}

\begin{figure*}
    \centering
  \includegraphics[width=0.9\textwidth]{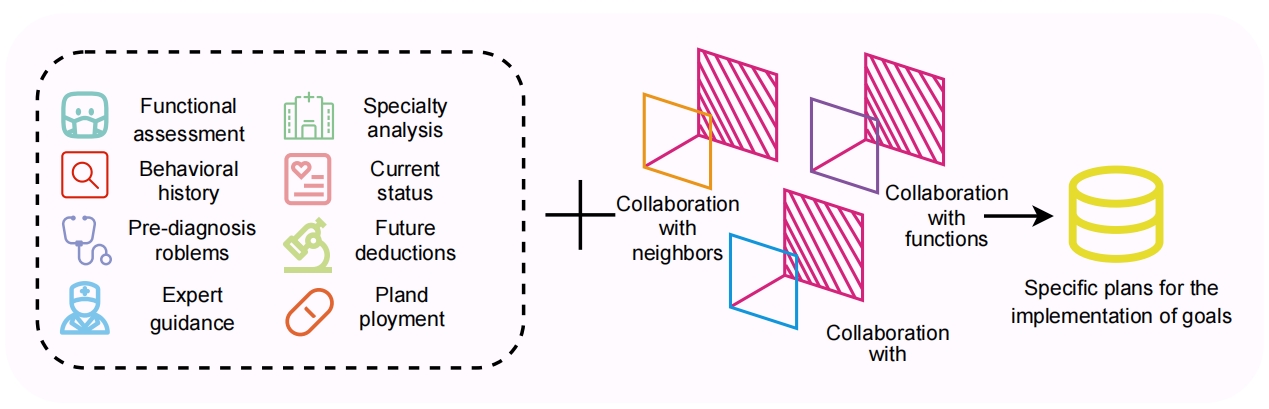} 
    \caption{Framework of the proposed \method Grounding Module that contextualizes symbol embeddings using the current state, intentions, and feedback signals.}
    \label{P2}
    \vspace{-3ex}
\end{figure*}

\begin{tcolorbox}[colback=lightgray!20,boxrule=1pt,sharp corners,fontupper=\footnotesize]
\textbf{\textit{Grounding Module (Figure \ref{P2}):}}
\begin{enumerate}
\item Contextualize the abstract traffic concepts and
symbols into grounded representations.
\item Map entities like intersections, vehicles, and
signal phases to their physical counterparts.
\item Resolve ambiguities and uncertainties in
grounding based on the current traffic context.
\item Adjust grounding strategies based on
feedback from execution agents and emerging
coordination patterns.
\item Provide grounded embeddings to inform the
execution agents' decision-making.
\end{enumerate}
\end{tcolorbox}

\begin{figure*}
    \centering
  \includegraphics[width=0.9\textwidth]{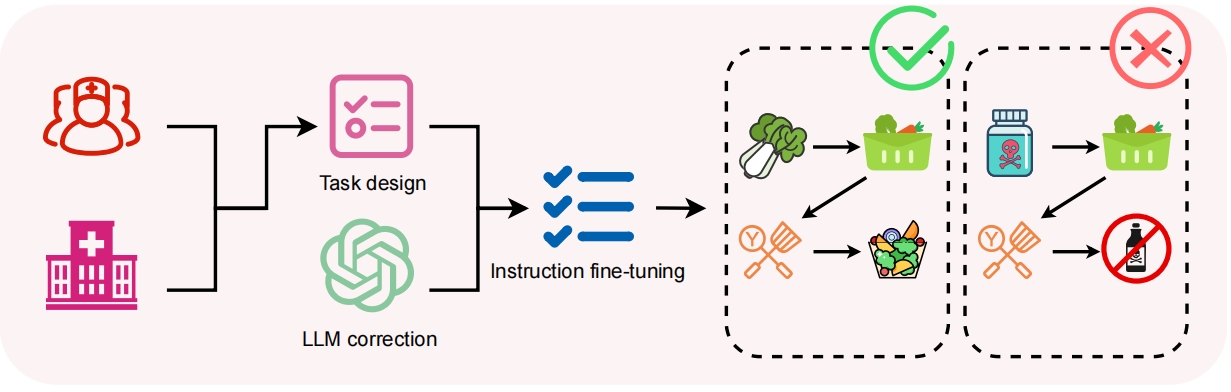} 
    \caption{Overview of our \method Cooperative Execution Module consisting of specialized agents that collaboratively execute actions and propagate intentions.}
    \label{P3}
    \vspace{-3ex}
\end{figure*}

\begin{figure*}
    \centering
  \includegraphics[width=0.9\textwidth]{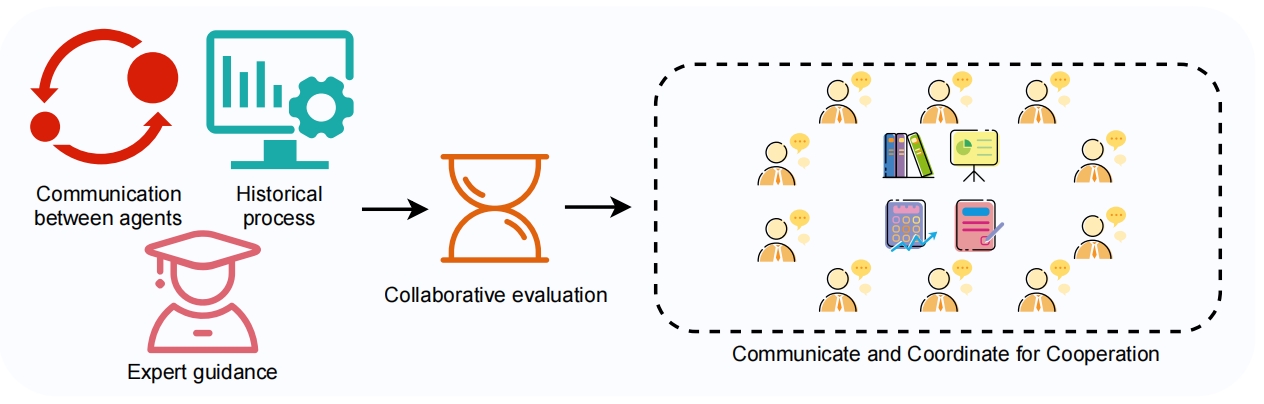} 
    \caption{Overview of the collaborative evaluation setup in the proposed \method framework.}
    \label{P4}
    \vspace{-3ex}
\end{figure*}

\begin{tcolorbox}[colback=lightgray!20,boxrule=1pt,sharp corners,fontupper=\footnotesize]
\textbf{\textit{Execution Module (Figure \ref{P3},\ref{P4}):}}
\begin{enumerate}
\item  Specialized agents monitor their respective
domains (vehicle counts, road conditions, signal
timings, etc.).
\item Agents communicate their local intentions and
goals to relevant teammates.
\item Agents align their actions based on shared
intentions and the coordinated plans.
\item Agents execute their assigned subtasks
(adjusting signal phases, routing emergency
vehicles, etc.).
\item Agents observe the impact of their actions and
provide feedback on emerging coordination
patterns.
\item Agents adapt their strategies dynamically
based on the feedback and changing traffic
conditions.
\item Agents continuously monitor and respond to
fluctuations in vehicle arrival rates and traffic
patterns.
\item Agents collaborate and coordinate their efforts
to collectively alleviate congestion and optimize
traffic flow.
\end{enumerate}
\end{tcolorbox}


\end{document}